\documentclass[11pt,a4paper]{article}
\usepackage[hyperref]{naaclhlt2019}
\usepackage{latexsym}
\usepackage{times}  
\usepackage{helvet}  
\usepackage{courier}  
\usepackage{url}  
\usepackage{graphicx}  
\usepackage{multirow}
\usepackage{soul}
\usepackage{arydshln}
\usepackage{amsmath}
\usepackage{amsfonts}
\usepackage{amssymb}
\usepackage{color}
\usepackage{graphicx}  
\usepackage{url}

\def\ignore#1{}

\aclfinalcopy 

\title{Predicting Annotation Difficulty to Improve Task Routing and Model Performance for Biomedical Information Extraction}

\author{Yinfei Yang \\
  Google AI \\
  {\tt yinfeiy@google.com} \\\And
  Oshin Agarwal \\
  University of Pennsylvania \\
  {\tt oagarwal@seas.upenn.edu} \\\AND
  Chris Tar \\
  Google AI \\
  {\tt ctar@google.com} \\ \And
    Byron C. Wallace \\
  Northeastern University \\
  {\tt b.wallace@northeastern.edu} \\\AND
  Ani Nenkova \\
  University of Pennsylvania \\
  {\tt nenkova@seas.upenn.edu} \\
  }

\date{}

\begin{document}
\maketitle
\begin{abstract}

Modern NLP systems require high-quality annotated data.
In specialized domains, expert annotations may be prohibitively expensive.
An alternative is to rely on crowdsourcing to reduce costs at the risk of introducing noise. In this paper we demonstrate that directly modeling instance difficulty can be used to improve model performance, and to route instances to appropriate annotators. 
Our difficulty prediction model combines two learned representations: a `universal' encoder trained on out-of-domain data, and a task-specific encoder. Experiments on a complex biomedical information extraction task using expert and lay annotators show that: (i) simply excluding from the training data instances predicted to be difficult yields a small boost in performance; (ii) using difficulty scores to weight instances during training provides further, consistent gains; (iii) assigning instances predicted to be difficult to domain experts is an effective strategy for task routing. Our experiments confirm the expectation that for specialized tasks expert annotations are higher quality than crowd labels, and hence preferable to obtain if practical. Moreover, augmenting small amounts of expert data with a larger set of lay annotations leads to further improvements in model performance. 

\end{abstract}

\section{Introduction}
Assembling training corpora of annotated natural language examples in specialized domains such as biomedicine poses considerable challenges.
Experts with the requisite domain knowledge to perform high-quality annotation tend to be expensive, while lay annotators may not have the necessary knowledge to provide high-quality annotations.
A practical approach for collecting a sufficiently large corpus would be to use crowdsourcing platforms like Amazon Mechanical Turk (MTurk). However, crowd workers in general are likely to provide noisy annotations ~\cite{DBLP:conf/starsem/AbadM16,DBLP:conf/eacl/PlankHS14,DBLP:conf/emnlp/AlonsoJLA15}, an issue exacerbated by the technical nature of specialized content. 
Some of this noise may reflect worker quality and can be modeled \cite{DBLP:conf/starsem/AbadM16,DBLP:conf/eacl/PlankHS14,DBLP:conf/acl/CohnS13,nguyen2017aggregating}, but for some instances lay people may simply lack the domain knowledge to provide useful annotation. 

In this paper we report experiments on the EBM-NLP corpus comprising crowdsourced annotations of medical literature \cite{nye-ACL-18}. We operationalize the concept of annotation difficulty and show how it can be exploited during training to improve information extraction models. We then obtain expert annotations for the abstracts predicted to be most difficult, as well as for a similar number of randomly selected abstracts. 
The annotation of highly specialized data and the use of lay and expert annotators allow us to examine the following key questions related to lay and expert annotations in specialized domains:

    {\em Can we predict item difficulty?} We define a training instance as difficult if a lay annotator or an automated model disagree on its labeling. We show that difficulty can be predicted, and that it is distinct from inter-annotator agreement. Further, such predictions can be used during training to improve information extraction models.  

    {\em Are there systematic differences between expert and lay annotations?} We observe decidedly lower agreement between lay workers as compared to domain experts. Lay annotations have high precision but low recall with respect to expert annotations in the new data that we collected. More generally, we expect lay annotations to be lower quality, which may translate to lower precision, recall, or both, compared to expert annotations. 
    
    {\em Can one rely solely on lay annotations?} Reasonable models can be trained using lay annotations alone, but similar performance can be achieved using markedly less expert data. This suggests that the optimal ratio of expert to crowd annotations for specialized tasks will depend on the cost and availability of domain experts.
    Expert annotations are preferable whenever its collection is practical. But in real-world settings, a combination of expert and lay annotations is better than using lay data alone.
    
    {\em Does it matter what data is annotated by experts?} We demonstrate that a system trained on combined data achieves better predictive performance when experts annotate difficult examples rather than instances selected at i.i.d. random.

Our contributions in this work are summarized as follows. We define a task difficulty prediction task and show how this is related to, but distinct from, inter-worker agreement. We introduce a new model for difficulty prediction combining learned representations induced via a pre-trained `universal' sentence encoder \cite{unec}, and a sentence encoder learned from scratch for this task. We show that predicting annotation difficulty can be used to improve the task routing and model performance for a biomedical information extraction task. 
Our results open up a new direction for ensuring corpus quality.
We believe that item difficulty prediction will likely be useful in other, non-specialized tasks as well, and that the most effective data collection in specialized domains requires research addressing the fundamental questions we examine here.

\section{Related Work}

Crowdsourcing annotation is now a well-studied problem ~\cite{snow2008cheap,DBLP:conf/starsem/AbadM16,DBLP:conf/eacl/PlankHS14,DBLP:conf/emnlp/AlonsoJLA15}. Due to the noise inherent in such annotations, there have also been considerable efforts to develop aggregation models that  minimize noise \cite{DBLP:conf/starsem/AbadM16,DBLP:conf/eacl/PlankHS14,DBLP:conf/acl/CohnS13,nguyen2017aggregating}.

There are also several surveys of crowdsourcing in biomedicine specifically \cite{good2013crowdsourcing,khare2015crowdsourcing,lee2017application}.
Some work in this space has contrasted model performance achieved using expert vs. crowd annotated training data \cite{zhai2013web,DBLP:journals/jbi/CocosQCM17,Dumitrache:2018:CGT:3232718.3152889}.  
Dumitrache \emph{et al.} \shortcite{Dumitrache:2018:CGT:3232718.3152889} concluded that performance is similar under these supervision types, finding no clear advantage from using expert annotators.  
This differs from our findings, perhaps owing to differences in design. The experts we used already hold advanced medical degrees, for instance, while those in prior work were medical students. Furthermore, the task considered here would appear to be of greater difficulty: even a system trained on $\sim$5k instances performs reasonably, but far from perfect. By contrast, in some of the prior work where experts and crowd annotations were deemed equivalent, a classifier trained on 300 examples can achieve very high accuracy \cite{DBLP:journals/jbi/CocosQCM17}. 

More relevant to this paper, prior work has investigated methods for `task routing' in active learning scenarios in which supervision is provided by heterogeneous labelers with varying levels of expertise \cite{yan2011active,donmez2008proactive,nguyen2015combining,wallace2011should,yan2011active}. The related question of whether effort is better spent collecting additional annotations for already labeled (but potentially noisily so) examples or novel instances has also been addressed \cite{sheng2008get}. What distinguishes the work here is our focus on providing an operational definition of instance \emph{difficulty}, showing that this can be predicted, and then using this to inform task routing.

\begin{table*}[!htbp]
\small
\centering
\begin{tabular}{|p{15.5cm}|}
\hline
\multicolumn{1}{|c|}{\textbf{Difficult Sentences}} \\ \hline
\textit{\textbf{[Population]}} \\
\textit{1. \ul{Primary RP}  were screened and assigned to either the nifedipine SR group (Group N) or the Ginkgo biloba extract group (Group G) in the ratio of 2:1 .} \\
\textit{2. A positive correlation was found for all methods in the  \ul{controls}  (r=0.83-0.94) and  \ul{RA patients} (r=0.51-0.69).} \\
\\
\hdashline

\textit{\textbf{[Interventions/Comparators]}} \\
\textit{1. They were all enrolled in  \ul{mainstream compulsory education}.} \\
\textit{2. \ul{RA} patients reported that they were less sedentary and engaged in more higher intensity  \ul{PA}  than what was objectively assessed.} \\
\\
\hdashline

\textit{\textbf{[Outcomes]}} \\
\textit{1. To develop a  \ul{cycle - based risk prediction model for neutropenic complications (NC)}  during chemotherapy with doxorubicin (DOX) or a pegy lated liposomal formulation (PLD) for patients with metastatic breast cancer (MBC).} \\
\textit{2. The purpose of this study was to evaluate the \ul{long-term safety and efficacy}  of risperidone.} \\
\\
\hline
\hline

\multicolumn{1}{|c|}{\textbf{Easy Sentences}} \\ \hline
\textit{\textbf{[Population]}} \\
\textit{1. A prospective study in \ul{80 patients}  was carried out.} \\
\textit{2. We studied \ul{200 women aged 35 years and older who had a family history of breast cancer in a first-degree relative.}} \\
\\
\hdashline

\textit{\textbf{[Interventions/Comparators]}} \\
\textit{1. \ul{Hormonal contraceptives} are used widely but their effects on HIV-1 risk are unclear.} \\
\textit{2. In the second group,  \ul{custom - fit MRI - based pin guides}  were used.} \\
\\
\hdashline

\textit{\textbf{[Outcomes]}} \\
\textit{1. \ul{Extrapyramidal AEs} were reported in 6(8\%) patients.} \\
\textit{2. The \ul{overall progression - free survival rates} were similar between the two arms (P=.095).} \\
\\
\hline
\end{tabular}
\caption{Example sentences are difficult or easy to annotate for crowd workers. The \ul{underlined text} are reference annotations from domain experts.}
\label{tab:example}
\end{table*}

\section{Application Domain}
\label{sec:corpus}
Our specific application concerns annotating abstracts of articles that describe the conduct and results of randomized controlled trials (RCTs). Experimentation in this domain has become easy with the recent release of the EBM-NLP~\cite{nye-ACL-18} corpus, which includes a reasonably large training dataset annotated via crowdsourcing, and a modest test set labeled by individuals with advanced medical training. 
More specifically, the training set comprises 4,741 medical article abstracts with crowdsourced annotations indicating snippets (sequences) that describe the Participants ({\sc p}), Interventions ({\sc i}), and Outcome ({\sc o}) elements of the respective RCT, and the test set is composed of 191 abstracts with {\sc p}, {\sc i}, {\sc o} sequence annotations from three medical experts.

Table \ref{tab:example} shows an example of difficult and easy examples according to our definition of difficulty.
The underlined text demarcates the (consensus) reference label provided by domain experts.
In the difficult examples, crowd workers marked text distinct from these reference annotations; whereas in the easy cases they reproduced them with reasonable fidelity.
The difficult sentences usually exhibit complicated structure and feature jargon.

An abstract may contain some `easy' and some `difficult' sentences. We thus perform our analysis at the sentence level.
We split abstracts into sentences using spaCy.\footnote{\url{https://spacy.io/}}
We excluded sentences that comprise fewer than two tokens, as these are likely an artifact of errors in sentence splitting.
In total, this resulted in 57,505 and 2,428 sentences in the train and test set abstracts, respectively.

\section{Quantifying Task Difficulty}

The test set includes annotations from both crowd workers and domain experts. We treat the latter as ground truth and then define the difficulty of sentences in terms of the observed agreement between expert and lay annotators.  
Formally, for annotation task $t$ and instance $i$:
\begin{equation}
   \text{Difficulty}_{ti} = \frac{\sum_{j=1}^n{f(\text{label}_{ij}, y_i})}{n}
\end{equation}
where $f$ is a scoring function that measures the quality of the label from worker $j$ for sentence $i$, as compared to a ground truth annotation, $y_i$. The difficulty score of sentence $i$ is taken as an average over the scores for all $n$ layworkers. 
We use Spearmans' correlation coefficient as a scoring function.
Specifically, for each sentence we create two vectors comprising counts of how many times each token was annotated by crowd and expert workers, respectively, and calculate the correlation between these. 
Sentences with no labels are treated as maximally easy; those with \emph{only} either crowd worker or expert label(s) are assumed maximally difficult.

The training set contains only crowdsourced annotations.
To label the training data, we use a 10-fold validation like setting. We iteratively retrain the LSTM-CRF-Pattern sequence tagger of Patel \emph{et al.} \shortcite{patel2018syntactic} on 9 folds of the training data and use that trained model to predict labels for the 10th. In this way we obtain predictions on the full training set.
We then use predicted spans as proxy `ground truth' annotations to calculate the difficulty score of sentences as described above; we normalize these to the [$0, 1$] interval.
We validate this approximation by comparing the proxy scores against reference scores over the test set, the Pearson's correlation coefficients are 0.57 for Population, 0.71 for Intervention and 0.68 for Outcome. 

There exist many sentences that contain neither manual nor predicted annotations.
We treat these as maximally easy sentences (with difficulty scores of 0).
Such sentences comprise 51\%, 42\% and 36\% for Population, Interventions and Outcomes data respectively, indicating that it is easier to identify sentences that have no Population spans, but harder to identify sentences that have no Interventions or Outcomes spans. This is intuitive as descriptions of the latter two tend to be more technical and dense with medical jargon.

\begin{figure}
    \centering
    \includegraphics[width=0.45\textwidth]{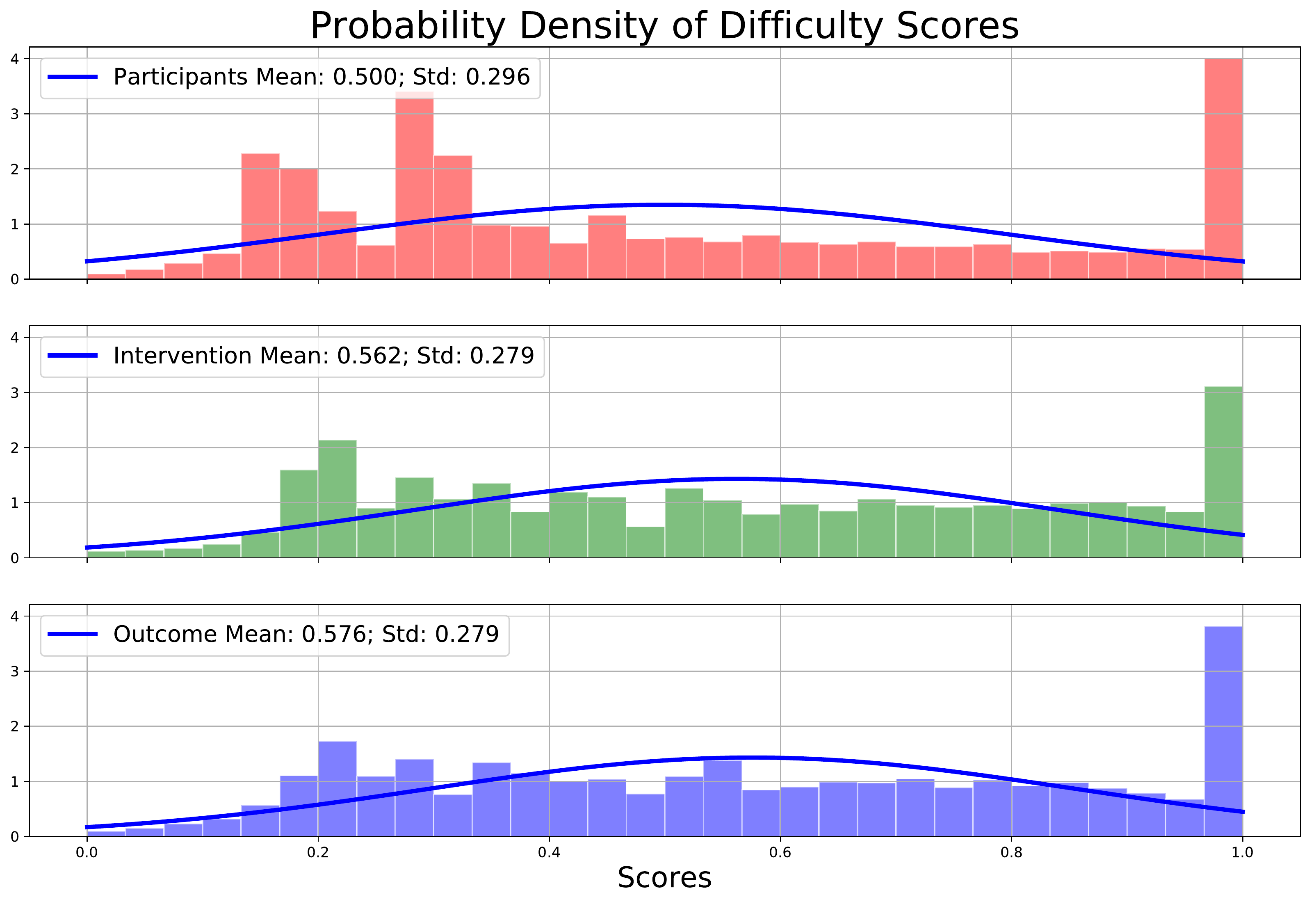}  
    \caption{Distributions of difficulty scores over all sentences that contain any span annotations} 
    \vspace{-.75em}
    \label{fig:difficulty_dist}
\end{figure}

We show the distribution of the automatically labeled scores for sentences that do contain spans in Figure \ref{fig:difficulty_dist}.
The mean of the Population ({\sc p}) sentence scores is significantly lower than that for other types of sentences ({\sc i} and {\sc o}), again indicating that they are easier on average to annotate.
This aligns with a previous finding that annotating Interventions and Outcomes is more difficult than annotating Participants \cite{nye-ACL-18}.

Many sentences contain spans tagged by the LSTM-CRF-Pattern model, but missed by all crowd workers, resulting in a maximally difficult score (1).
Inspection of such sentences revealed that some are truly difficult examples, but others are tagging model errors. 
In either case, such sentences have confused workers and/or the model, and so we retain them all as `difficult' sentences.

Content describing the {\sc p}, {\sc i} and {\sc o}, respectively, is quite different. As such, one sentence usually contains (at most) only one of these three content types.
We thus treat difficulty prediction for the respective label types as separate tasks.

\section{Difficulty is not Worker Agreement}

Our definition of difficulty is derived from agreement between expert and crowd annotations for the test data, and agreement between a predictive model and crowd annotations in the training data. It is reasonable to ask if these measures are related to inter-annotator agreement, a metric often used in language technology research to identify ambiguous or difficult items. Here we explicitly verify that our definition of difficulty only weakly correlates with inter-annotator agreement.  

We calculate inter-worker agreement between crowd and expert annotators using Spearman's correlation coefficient.
As shown in Table \ref{tab:inter_agreement}, average agreement between domain experts are considerably higher than agreements between crowd workers for all three label types. This is a clear indication that the crowd annotations are noisier.

Furthermore, we compare the correlation between inter-annotator agreement and difficulty scores in the training data.
Given that the majority of sentences do not contain a PICO span, we only include in these calculations those that contain a reference label.
Pearson's r are 0.34, 0.30 and 0.31 for {\sc p}, {\sc i} and {\sc o}, respectively, confirming that inter-worker agreement and our proposed difficulty score are quite distinct.

\begin{table}
\centering
\begin{tabular}{c c c c}
\hline \bf Workers & \bf P & \bf I & \bf O \\ \hline
crowd workers & 0.52 &  0.43 &  0.41 \\
domain experts & 0.74 &  0.68 &  0.57 \\
\hline
\end{tabular} 
\caption{Average inter-worker agreement.}
\label{tab:inter_agreement}
\end{table}

\section{Predicting Annotation Difficulty}

We treat difficulty prediction as a regression problem, and propose and evaluate neural model variants for the task.
We first train RNN~\cite{rnn} and CNN~\cite{cnn} models. 

\begin{figure}
    \centering
    \includegraphics[width=0.375\textwidth]{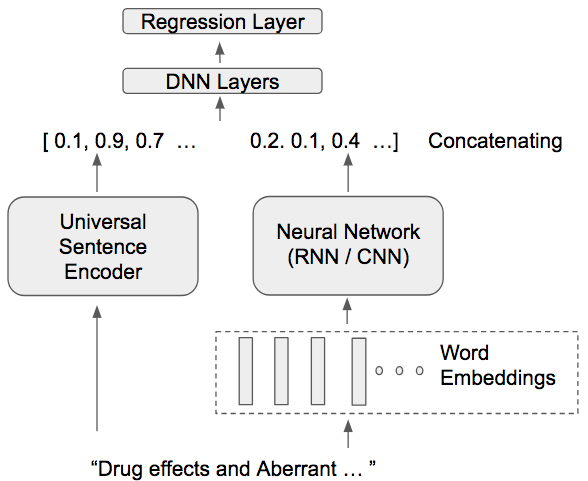}
    \caption{Model architecture.}
    \label{fig:model}
\end{figure}

We also use the \emph{universal sentence encoder} (USE)~\cite{unec} to induce sentence representations, and train a model using these as features. Following \cite{unec}, we then experiment with an ensemble model that combines the `universal' and task-specific representations to predict annotation difficulty.
We expect these universal embeddings to capture general, high-level semantics, and the task specific representations to capture more granular information. Figure \ref{fig:model} depicts the model architecture. Sentences are fed into both the universal sentence encoder and, separately, a task specific neural encoder, yielding two representations. We concatenate these and pass the combined vector to the regression layer.

\subsection{Experimental Setup and Results}

We trained models for each label type separately.
Word embeddings were initialized to 300d GloVe vectors~\cite{pennington2014glove} trained on common crawl data;\footnote{\url{http://nlp.stanford.edu/data/glove.840B.300d.zip}} 
these are fine-tuned during training.
We used the Adam optimizer \cite{kingma2014adam} with learning rate and decay set to 0.001 and 0.99, respectively. We used batch sizes of 16.

We used the large version of the universal sentence encoder\footnote{\url{https://tfhub.dev/google/universal-sentence-encoder-large/3}} with a transformer \cite{transformer}.
We did not update the pretrained sentence encoder parameters during training.
All hyperparamaters for all models (including hidden layers, hidden sizes, and dropout) were tuned using Vizier \cite{golovin2017} via 10-fold cross validation on the training set maximizing for F1.\footnote{This performs random search over the number of hidden layers (1-5), hidden sizes (128-1024), and dropout (0.1- 0.5).}

As a baseline, we also trained a linear Support-Vector Regression \cite{liblinear} model on $n$-gram features ($n$ ranges from 1 to 3).\footnote{We perform gridsearch over the hyperparameter \textit{C}.}

Table \ref{tab:res_difficulty} reports Pearson correlation coefficients between the predictions with each of the neural models and the ground truth difficulty scores.
Rows 1-4 correspond to individual models, and row 5 reports the ensemble performance.
Columns correspond to label type. 
Results from all models outperform the baseline SVR model: Pearson's correlation coefficients range from 0.550 to 0.622. The regression correlations are the lowest.

The RNN model realizes the strongest performance among the stand-alone (non-ensemble) models, outperforming variants that exploit CNN and USE representations.
Combining the RNN and USE further improves results. 
We hypothesize that this is due to complementary sentence information encoded in universal representations.

For all models, correlations for Intervention and Outcomes are higher than for Population, which is expected given the difficulty distributions in Figure \ref{fig:difficulty_dist}. In these, the sentences are more uniformly distributed, with a fair number of difficult and easier sentences. By contrast, in Population there are a greater number of easy sentences and considerably fewer difficult sentences, which makes the difficulty ranking task particularly challenging.

\begin{table}
    \centering
    \begin{tabular}{ c c c c }
        \hline
         & \bf{P} & \bf{I} & \bf{O}  \\ \hline
        NGRAM+SVR & 0.455 & 0.311 & 0.541 \\
        RNN     & 0.521 & 0.555 & 0.601 \\
        CNN     & 0.470 & 0.522 & 0.550 \\
        USE    & 0.492 & 0.518 & 0.580 \\
        \hline
        USE+RNN  & 0.550 & 0.604 & 0.622 \\ 
        \hline
    \end{tabular}
     \caption{Pearson correlation coefficients of sentence difficulty predictions.}
    \label{tab:res_difficulty}
\end{table}

\section{Better IE with Difficulty Prediction}

\begin{table*}
    \centering
    \begin{tabular}{c |c c c | c c c | c c c}
    \hline
        \multirow{2}{*}{\bf Model} &  \multicolumn{3}{c|}{\textbf{Precision} } &  \multicolumn{3}{c|}{\textbf{Recall} } &  \multicolumn{3}{c}{\textbf{F1} } \\
        \cline{2-10}
          & \bf P & \bf I & \bf O & \bf P & \bf I & \bf O & \bf P & \bf I & \bf O \\ 
        \hline
        Base model                  & {\it 81.54} & 81.99         & 78.01 &       64.22 & 54.19 & 54.84 & 71.85 & 65.25 & 64.40 \\
        Re-weight by agreement      & 78.63       & {\it 82.24}   & {\it 82.19} & 66.47 & 54.45 & 55.22 & 72.04 & 65.53 & 66.06 \\
        Re-weight by difficulty     & 79.57       & 74.69         & 73.92        & {\it 70.31} & {\it 63.71} & {\it 64.96} & {\bf 74.65} & {\bf 68.76} & {\bf 69.15} \\
        \hline
    \end{tabular}
        \caption{Medical IE performance by re-weighting sentences according to predicted agreement or difficulty scores. }
    \label{tab:res_reweight}
\end{table*}

We next present experiments in which we attempt to use the predicted difficulty during training to improve models for information extraction of descriptions of Population, Interventions and Outcomes from medical article abstracts. We investigate two uses: (1) simply removing the most difficult sentences from the training set, and, (2) re-weighting the most difficult sentences.

We again use LSTM-CRF-Pattern as the base model and experimenting on the EBM-NLP corpus~\cite{nye-ACL-18}. This is trained on either (1) the training set with difficult sentences removed, or (2) the full training set but with instances re-weighted in proportion to their predicted difficulty score.
Following \cite{nye-ACL-18}, we use the Adam optimizer with learning rate of 0.001, decay 0.9, batch size 20 and dropout 0.5.
We use pretrained 200d GloVe vectors~\cite{pennington2014glove}\footnote{\url{http://nlp.stanford.edu/data/glove.6B.zip}} to initialize word embeddings, and use 100d hidden char representations.
Each word is thus represented with 300 dimensions in total. 
The hidden size is 100 for the LSTM in the character representation component, and 200 for the LSTM in the information extraction component.
We train for 15 epochs, saving parameters that achieve the best F1 score on a nested development set.

\subsection{Removing Difficult Examples}

We first evaluate changes in performance induced by training the sequence labeling model using less data by removing difficult sentences prior to training. The hypothesis here is that these difficult instances are likely to introduce more noise than signal.
We used a cross-fold approach to predict sentence difficulties, training on 9/10ths of the data and scoring the remaining 1/10th at a time. We then sorted sentences by predicted difficulty scores, and experimented with removing increasing numbers of these (in order of difficulty) prior to training the LSTM-CRF-Pattern model. 

\begin{figure}
    \centering
    \includegraphics[width=0.48\textwidth]{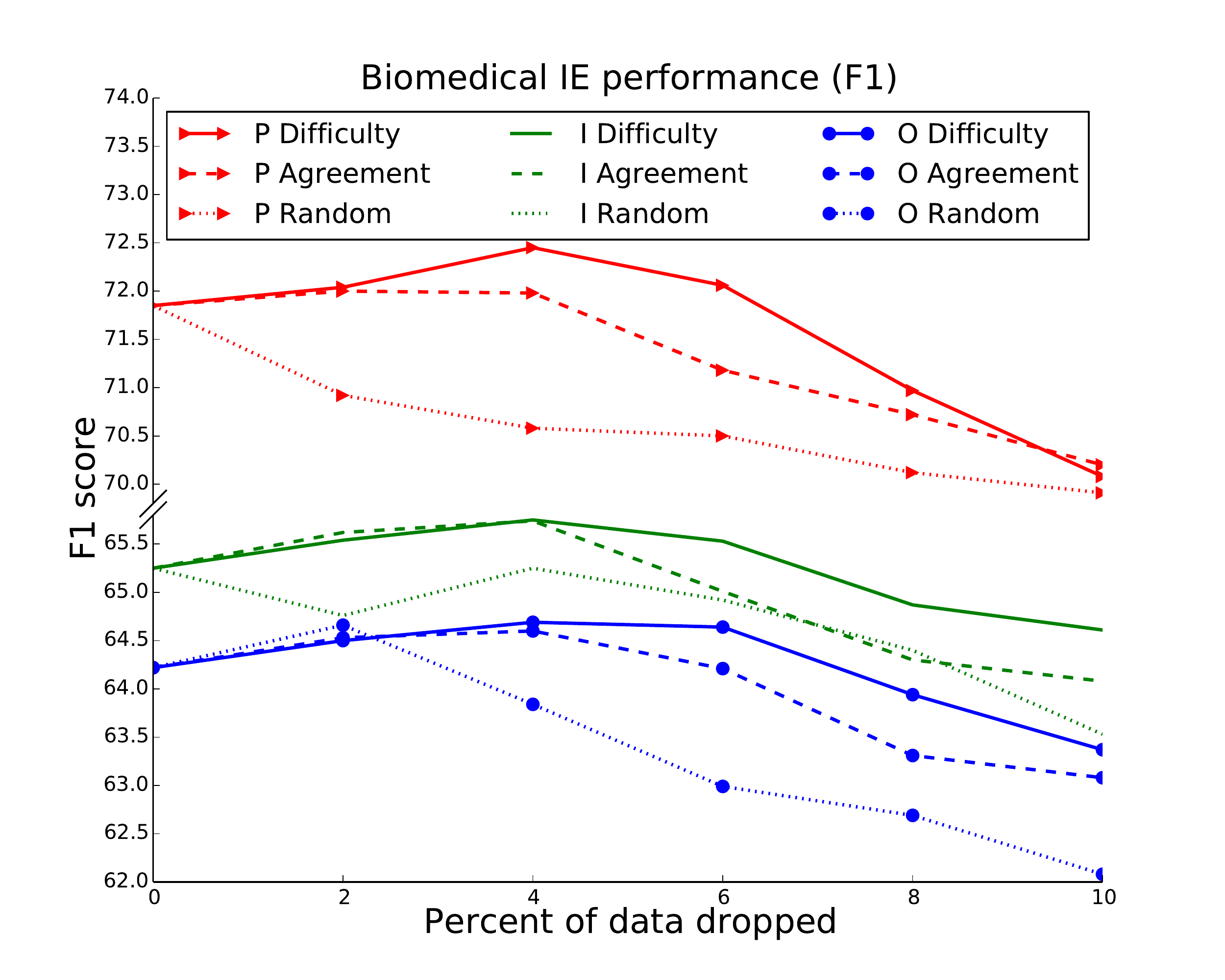}
    \caption{F1 scores achieved when removing increasingly large fractions of the training data.}
    \label{fig:f1_drop}
\end{figure}

Figure \ref{fig:f1_drop} shows the results achieved by the LSTM-CRF-Pattern model after discarding increasing amounts of the training data:
the $x$ and $y$ axes correspond to the the percentage of data removed and F1 scores, respectively.
We contrast removing sentences predicted to be difficult with removing them {\em (a)} randomly (i.i.d.), and, {\em (b)} in inverse order of predicted inter-annotator agreement.
The agreement prediction model is trained exactly the same like difficult prediction model, with simply changing the difficult score to annotation agreement.
F1 scores actually improve (marginally) when we remove the most difficult sentences, up until we drop 4\% of the data for Population and Interventions, and 6\% for Outcomes.
Removing training points at i.i.d. random degrades performance, as expected.
Removing sentences in order of disagreement seems to have similar effect as removing them by difficulty score when removing small amount of the data, but the F1 scores drop much faster when removing more data.
These findings indicate that sentences predicted to be difficult are indeed noisy, to the extent that they do not seem to provide the model useful signal.

\subsection{Re-weighting by Difficulty}

We showed above that removing a small number of the most difficult sentences does not harm, and in fact modestly improves, medical IE model performance.
However, using the available data we are unable to test if this will be useful in practice, as we would need additional data to determine how many difficult sentences should be dropped.

We instead explore an alternative, practical means of exploiting difficulty predictions: we re-weight sentences during training inversely to their \emph{predicted} difficulty.
Formally, we weight sentence $i$ with difficulty scores above $\tau$ according to: $1-a\cdot(d_i-\tau)/(1-\tau)$, where $d_i$ is the difficulty score for sentence $i$, and $a$ is a parameter codifying the minimum weight value.
We set $\tau$ to 0.8 so as to only re-weight sentences with difficulty in the top 20th percentile,
and we set $a$ to 0.5. The re-weighting is equivalent to down-sampling the difficult sentences. LSTM-CRF-Pattern is our base model.

Table \ref{tab:res_reweight} reports the precision, recall and F1 achieved both with and without sentence re-weighting.
Re-weighting improves all metrics modestly but consistently.
All F1 differences are statistically significant under a sign test ($p<0.01$).
The model with best precision is different for Patient, Intervention and Outcome labels. However re-weighting by difficulty does consistently yield the best recall for all three extraction types, with the most notable improvement for {\sc i } and {\sc o}, where recall improved by 10 percentage points. This performance increase translated to improvements in F1 across all types, as compared to the base model and to re-weighting by agreement.

\section{Involving Expert Annotators}

The preceding experiments demonstrate that re-weighting difficult sentences annotated by the crowd generally improves the extraction models. 
Presumably the performance is influenced by the annotation quality. 

We now examine the possibility that the higher quality and more consistent annotations of domain experts on the difficult instances will benefit the extraction model. This simulates an annotation strategy in which we route difficult instances to domain experts and easier ones to crowd annotators. We also contrast the value of difficult data to that of an i.i.d. random sample of the same size, both annotated by experts. 

\subsection{Expert annotations of Random and Difficult Instances}
We  re-annotate by experts a subset of most difficult instances and the same number of random instances.
As collecting annotations from experts is slow and expensive, we only re-annotate the difficult instances for the interventions extraction task. 
We re-annotate the abstracts which cover the sentences with predicted difficulty scores in the top 5 percentile.
We rank the abstracts from the training set by the count of difficult sentences, and re-annotate the abstracts that contain the most difficult sentences.
Constrained by time and budget, we select only 2000 abstracts for re-annotation; 1000 of these are top-ranked, and 1000 are randomly sampled. This re-annotation cost \$3,000. We have released the new annotation data at: \url{https://github.com/bepnye/EBM-NLP}.  

Following \cite{nye-ACL-18}, we recruited five medical experts via Up-work\footnote{https://www.upwork.com} with advanced medical training and strong technical reading/writing skills. 
The expert annotator were asked to read the entire abstract and highlight, using the BRAT toolkit~\cite{brat}, all spans describing medical Interventions.
Each abstract is only annotated by one expert. We examined 30 re-annotated abstracts to ensure the annotation quality before hiring the annotator.

Table \ref{tab:res_intervetion_reanno} presents the results of LSTM-CRF-Pattern model trained on the reannotated difficult subset and the random subset.
The first two rows show the results for models trained with expert annotations. The model trained on random data has a slightly better F1 than that trained on the same amount of difficult data. The model trained on random data has higher precision but lower recall.

Rows 3 and 4 list the results for models trained on the same data but with crowd annotation.
Models trained with expert-annotated data are clearly superior to those trained with crowd labels with respect to F1, indicating that the experts produced higher quality annotations. For crowdsourced annotations, training the model with data sampled at i.i.d. random achieves 2\% higher F1 than when difficult instances are used. When expert annotations are used, this difference is less than 1\%.
This trend in performance may be explained by differences in annotation quality: the randomly sampled set was more consistently annotated by both experts and crowd because the difficult set is harder. However, in both cases expert annotations are better, with a bigger difference between the expert and crowd models on the difficult set.

The last row is the model trained on all 5k abstracts with crowd annotations.
Its F1 score is lower than either expert model trained on only 20\% of data, suggesting that expert annotations should be collected whenever possible.
Again the crowd model on complete data has higher precision than expert models but its recall is much lower.

\begin{table}
    \centering
    \begin{tabular}{ c c c c }
        \hline
         & \bf{Precision} & \bf{Recall} & \bf{F1}  \\
         \hline
        Difficult-Expert    & 68.46 & 65.06 & 66.72 \\
        Random-Expert       & 70.84 & 63.46 & 67.04 \\
        \hline
        Difficult-Crowd     & 83.68 & 44.63 & 58.45 \\
        Random-Crowd        & 78.55 & 49.23 & 60.52 \\
        \hline
        Base (All-Crowd)    & 81.99 & 54.19 & 65.25 \\
        \hline
    \end{tabular}
        \caption{Interventions IE model performance trained crowd or expert. The first four models are trained with a subset of 1k abstracts and the base model is trained with all 5k abstracts.}
    \label{tab:res_intervetion_reanno}
\end{table}

\subsection{Routing To Experts or Crowd}

\begin{table}[!htb]
    \centering
    \begin{tabular}{ c c c c }
        \hline
         & \bf{Precision} & \bf{Recall} & \bf{F1}  \\ \hline
        (D)ifficult-Expert      & 68.46 & 65.06 & 66.72 \\
        (R)andom-Expert         & 70.84 & 63.46 & 67.04 \\
        D+R                   & 68.57 & 67.54 & 69.62 \\ \hline
        D+Other               & 83.55 & 60.73 & 70.33 \\
        R+Other               & 85.10 & 57.08 & 68.34 \\
        D+R+Other           & 82.19 & 61.88 & 70.60 \\
        \hline
    \end{tabular}
        \caption{Interventions IE model performance trained by mixing annotations from experts and crowd workers. [D]: Difficult-Expert; [R]: Random-Expert; [Other]: the rest of the abstracts with crowd annotation only.}
    \label{tab:res_intervetion_combine}
\end{table}

So far a system was trained on one type of data, either labeled by crowd or experts.
We now examine the performance of a system trained on data that was routed to either experts or crowd annotators depending on their predicted difficult. Given the results presented so far mixing annotators may be beneficial given their respective trade-offs of precision and recall. 
We use the annotations from experts for an abstract if it exists otherwise use crowd annotations. The results are presented in Table \ref{tab:res_intervetion_combine}.

Rows 1 and 2 repeat the performance of the models trained on difficult subset and random subset with expert annotations only respectively.
The third row is the model trained by combining difficult and random subsets with expert annotations.
There are around 250 abstracts in the overlap of these two sets, so there are total 1.75k abstracts used for training the D+R model.
Rows 4 to 6 are the models trained on all 5k abstracts with mixed annotations, where \textit{Other} means the  rest  of  the  abstracts  with  crowd annotation only.

The results show adding more training data with crowd annotation still improves at least 1 point F1 score in all three extraction tasks.
The improvement when the difficult subset with expert annotations is mixed with the remaining crowd annotation is 3.5 F1 score, much larger than when a random set of expert annotations are added. 
The model trained with re-annotating the difficult subset (D+Other) also outperforms the model with re-annotating the random subset (R+Other) by 2 points in F1. 
The model trained with re-annotating both of difficult and random subsets (D+R+Other), however, achieves only marginally higher F1 than the model trained with the re-annotated difficult subset (D+Other).

In sum, the results clearly indicate that mixing expert and crowd annotations leads to better models than using solely crowd data, and better than using expert data alone. More importantly, there is greater gain in performance when instances are routed according to difficulty, as compared to randomly selecting the data for expert annotators. These findings align with our motivating hypothesis that annotation quality for difficult instances is important for final model performance.
They also indicate that mixing annotations from expert and crowd could be an effective way to achieve acceptable model performance given a limited budget.

\subsection{How Many Expert Annotations?}

\begin{figure}
    \centering
    \includegraphics[width=0.475\textwidth]{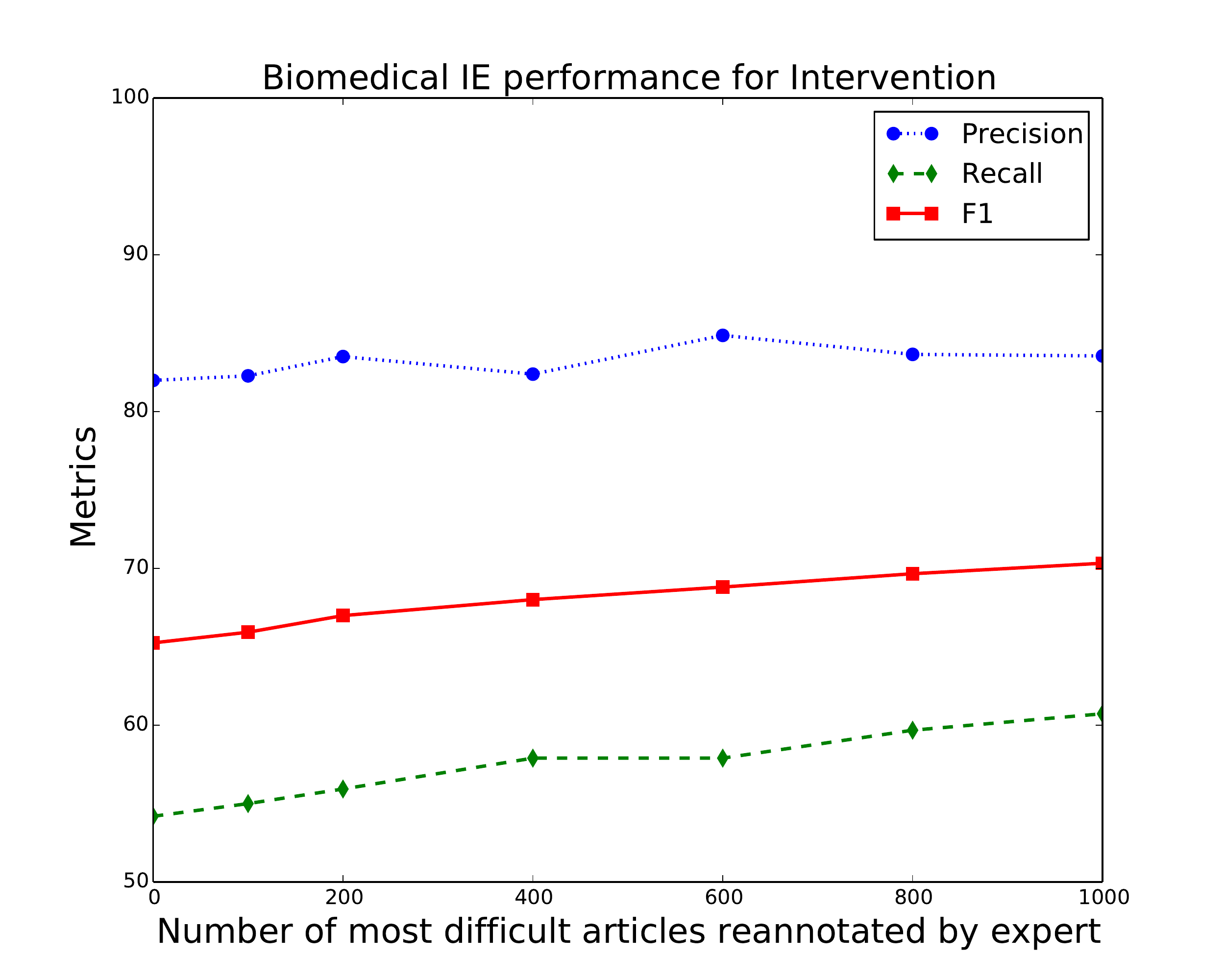}
    \caption{Precision/Recall/F1 as a function of the number of articles re-annotated by expert, in decreasing order of difficulty.}
    \label{fig:metric_mixing}
\end{figure}

We established that crowd annotation are still useful in supplementing expert annotations for medical IE. Obtaining expert annotations for the one thousand most difficult instances greatly improved the model performance. However the choice of how many difficult instances to annotate was an uninformed choice. Here we check if less expert data would have yielded similar gains. Future work will need to address how best to choose this parameter for a routing system.

We simulate a routing scenario in which we send consecutive batches of the most difficult examples to the experts for annotation. We track changes in performance as we increase the number of most-difficult-articles sent to domain experts. As shown in Figure \ref{fig:metric_mixing}, adding expert annotations for difficult articles consistently increases F1 scores. The performance gain is mostly from increased recall; the precision changes only a bit with higher quality annotation. This observation implies that crowd workers often fail to mark target tokens, but do not tend to produce large numbers of false positives. We suspect such failures to identify relevant spans/tokens are due to insufficient domain knowledge possessed by crowd workers.

The F1 score achieved after re-annotating the 600 most-difficult articles reaches 68.1\%, which is close to the performance when re-annotating 1000 random articles. This demonstrates the effectiveness of recognizing difficult instances. The trend when we use up all expert data is still upward, so adding even more expert data is likely to further improve performance. Unfortunately we exhausted our budget and were not able to obtain additional expert annotations. It is likely that as the size of the expert annotations increases, the value of crowd annotations will diminish. This investigation is left for future work.

\section{Conclusions}

We have introduced the task of predicting annotation difficulty for biomedical information extraction (IE).
We trained neural models using different learned representations to score texts in terms of their difficulty. 
Results from all models were strong with Pearson’s correlation coefficients higher than 0.45 in almost all evaluations, indicating the feasibility of this task.
An ensemble model combining universal and task specific feature sentence vectors yielded the best results.

Experiments on biomedical IE tasks show that removing up to $\sim$10\% of the sentences predicted to be most difficult did not decrease model performance, and that re-weighting sentences inversely to their difficulty score during training improves predictive performance. Simulations in which difficult examples are routed to experts and other instances to crowd annotators yields the best results, outperforming the strategy of randomly selecting data for expert annotation, and substantially improving upon the approach of relying exclusively on crowd annotations. In future work, routing strategies based on instance difficulty could be further investigated for budget-quality trade-off. 

\section*{Acknowledgements}
This work has been partially supported by NSF1748771 grant. Wallace was support in part by NIH/NLM R01LM012086.

\bibliography{naaclhlt2019}

\begin{thebibliography}{28}
\expandafter\ifx\csname natexlab\endcsname\relax\def\natexlab#1{#1}\fi

\bibitem[{Abad and Moschitti(2016)}]{DBLP:conf/starsem/AbadM16}
Azad Abad and Alessandro Moschitti. 2016.
\newblock \href {http://aclweb.org/anthology/S/S16/S16-2018.pdf} {Taking the
  best from the crowd: Learning question passage classification from noisy
  data}.
\newblock In \emph{Proceedings of the Fifth Joint Conference on Lexical and
  Computational Semantics, *SEM@ACL 2016, Berlin, Germany, 11-12 August 2016}.

\bibitem[{Alonso et~al.(2015)Alonso, Johannsen, de~Lacalle, and
  Agirre}]{DBLP:conf/emnlp/AlonsoJLA15}
H{\'{e}}ctor~Mart{\'{\i}}nez Alonso, Anders Johannsen, Oier~Lopez de~Lacalle,
  and Eneko Agirre. 2015.
\newblock \href {https://doi.org/10.18653/v1/W15-2711} {Predicting word sense
  annotation agreement}.
\newblock In \emph{Proceedings of the First Workshop on Linking Computational
  Models of Lexical, Sentential and Discourse-level Semantics, LSDSem@EMNLP
  2015, Lisbon, Portugal, September 18, 2015}, pages 89--94.

\bibitem[{Cer et~al.(2018)Cer, Yang, Kong, Hua, Limtiaco, John, Constant,
  Guajardo-Cespedes, Yuan, Tar et~al.}]{unec}
Daniel Cer, Yinfei Yang, Sheng-yi Kong, Nan Hua, Nicole Limtiaco, Rhomni~St
  John, Noah Constant, Mario Guajardo-Cespedes, Steve Yuan, Chris Tar, et~al.
  2018.
\newblock Universal sentence encoder for english.
\newblock In \emph{Proceedings of the 2018 Conference on Empirical Methods in
  Natural Language Processing: System Demonstrations}, pages 169--174.

\bibitem[{Chung et~al.(2014)Chung, Gulcehre, Cho, and Bengio}]{rnn}
Junyoung Chung, Caglar Gulcehre, KyungHyun Cho, and Yoshua Bengio. 2014.
\newblock Empirical evaluation of gated recurrent neural networks on sequence
  modeling.
\newblock \emph{arXiv preprint arXiv:1412.3555}.

\bibitem[{Cocos et~al.(2017)Cocos, Qian, Callison{-}Burch, and
  Masino}]{DBLP:journals/jbi/CocosQCM17}
Anne Cocos, Ting Qian, Chris Callison{-}Burch, and Aaron~J. Masino. 2017.
\newblock \href {https://doi.org/10.1016/j.jbi.2017.04.003} {Crowd control:
  Effectively utilizing unscreened crowd workers for biomedical data
  annotation}.
\newblock \emph{Journal of Biomedical Informatics}, 69:86--92.

\bibitem[{Cohn and Specia(2013)}]{DBLP:conf/acl/CohnS13}
Trevor Cohn and Lucia Specia. 2013.
\newblock \href {http://aclweb.org/anthology/P/P13/P13-1004.pdf} {Modelling
  annotator bias with multi-task gaussian processes: An application to machine
  translation quality estimation}.
\newblock In \emph{Proceedings of the 51st Annual Meeting of the Association
  for Computational Linguistics, {ACL} 2013, 4-9 August 2013, Sofia, Bulgaria,
  Volume 1: Long Papers}, pages 32--42.

\bibitem[{Donmez and Carbonell(2008)}]{donmez2008proactive}
Pinar Donmez and Jaime~G Carbonell. 2008.
\newblock Proactive learning: cost-sensitive active learning with multiple
  imperfect oracles.
\newblock In \emph{Proceedings of the 17th ACM conference on Information and
  knowledge management}, pages 619--628. ACM.

\bibitem[{Dumitrache et~al.(2018)Dumitrache, Aroyo, and
  Welty}]{Dumitrache:2018:CGT:3232718.3152889}
Anca Dumitrache, Lora Aroyo, and Chris Welty. 2018.
\newblock \href {http://doi.acm.org/10.1145/3152889} {Crowdsourcing ground
  truth for medical relation extraction}.
\newblock \emph{ACM Trans. Interact. Intell. Syst.}, 8(2):11:1--11:20.

\bibitem[{Fan et~al.(2008)Fan, Chang, Hsieh, Wang, and Lin}]{liblinear}
Rong-En Fan, Kai-Wei Chang, Cho-Jui Hsieh, Xiang-Rui Wang, and Chih-Jen Lin.
  2008.
\newblock {LIBLINEAR}: A library for large linear classification.
\newblock \emph{Journal of Machine Learning Research}, 9:1871--1874.

\bibitem[{Golovin et~al.(2017)Golovin, Solnik, Moitra, Kochanski, Karro, and
  Sculley}]{golovin2017}
Daniel Golovin, Benjamin Solnik, Subhodeep Moitra, Greg Kochanski, John Karro,
  and D.~Sculley. 2017.
\newblock Google vizier: A service for black-box optimization.
\newblock In \emph{Proceedings of KDD '17}.

\bibitem[{Good and Su(2013)}]{good2013crowdsourcing}
Benjamin~M Good and Andrew~I Su. 2013.
\newblock Crowdsourcing for bioinformatics.
\newblock \emph{Bioinformatics}, 29(16):1925--1933.

\bibitem[{Khare et~al.(2015)Khare, Good, Leaman, Su, and
  Lu}]{khare2015crowdsourcing}
Ritu Khare, Benjamin~M Good, Robert Leaman, Andrew~I Su, and Zhiyong Lu. 2015.
\newblock Crowdsourcing in biomedicine: challenges and opportunities.
\newblock \emph{Briefings in bioinformatics}, 17(1):23--32.

\bibitem[{Kim(2014)}]{cnn}
Yoon Kim. 2014.
\newblock \href {http://www.aclweb.org/anthology/D14-1181} {Convolutional
  neural networks for sentence classification}.
\newblock In \emph{Proceedings of the 2014 Conference on Empirical Methods in
  Natural Language Processing (EMNLP)}, pages 1746--1751, Doha, Qatar.
  Association for Computational Linguistics.

\bibitem[{Kingma and Ba(2014)}]{kingma2014adam}
Diederik~P Kingma and Jimmy Ba. 2014.
\newblock Adam: A method for stochastic optimization.
\newblock \emph{arXiv preprint arXiv:1412.6980}.

\bibitem[{Lee et~al.(2017)Lee, Arida, and Donovan}]{lee2017application}
Young~Ji Lee, Janet~A Arida, and Heidi~S Donovan. 2017.
\newblock The application of crowdsourcing approaches to cancer research: a
  systematic review.
\newblock \emph{Cancer medicine}, 6(11):2595--2605.

\bibitem[{Nguyen et~al.(2017)Nguyen, Wallace, Li, Nenkova, and
  Lease}]{nguyen2017aggregating}
An~T Nguyen, Byron~C Wallace, Junyi~Jessy Li, Ani Nenkova, and Matthew Lease.
  2017.
\newblock Aggregating and predicting sequence labels from crowd annotations.
\newblock In \emph{Proceedings of the conference. Association for Computational
  Linguistics. Meeting}, volume 2017, page 299. NIH Public Access.

\bibitem[{Nguyen et~al.(2015)Nguyen, Wallace, and Lease}]{nguyen2015combining}
An~Thanh Nguyen, Byron~C Wallace, and Matthew Lease. 2015.
\newblock Combining crowd and expert labels using decision theoretic active
  learning.
\newblock In \emph{Third AAAI conference on human computation and
  crowdsourcing}.

\bibitem[{Nye et~al.(2018)Nye, Li, Patel, Yang, Marshall, Nenkova, and
  Wallace}]{nye-ACL-18}
Benjamin Nye, Jessy Li, Roma Patel, Yinfei Yang, Iain Marshall, Ani Nenkova,
  and Byron~C. Wallace. 2018.
\newblock {A Corpus with Multi-Level Annotations of Patients, Interventions and
  Outcomes to Support Language Processing for Medical Literature}.
\newblock In \emph{Proceedings of the Conference of the Association for
  Computational Linguistics (ACL)}.

\bibitem[{Patel et~al.(2018)Patel, Yang, Marshall, Nenkova, and
  Wallace}]{patel2018syntactic}
Roma Patel, Yinfei Yang, Iain Marshall, Ani Nenkova, and Byron Wallace. 2018.
\newblock Syntactic patterns improve information extraction for medical search.
\newblock In \emph{Proceedings of the Conference of the North American Chapter
  of the Association for Computational Linguistics (NAACL)}.

\bibitem[{Pennington et~al.(2014)Pennington, Socher, and
  Manning}]{pennington2014glove}
Jeffrey Pennington, Richard Socher, and Christopher~D. Manning. 2014.
\newblock \href {http://www.aclweb.org/anthology/D14-1162} {Glove: Global
  vectors for word representation}.
\newblock In \emph{Empirical Methods in Natural Language Processing (EMNLP)},
  pages 1532--1543.

\bibitem[{Plank et~al.(2014)Plank, Hovy, and
  S{\o}gaard}]{DBLP:conf/eacl/PlankHS14}
Barbara Plank, Dirk Hovy, and Anders S{\o}gaard. 2014.
\newblock Learning part-of-speech taggers with inter-annotator agreement loss.
\newblock In \emph{Proceedings of the 14th Conference of the European Chapter
  of the Association for Computational Linguistics, {EACL} 2014, April 26-30,
  2014, Gothenburg, Sweden}, pages 742--751.

\bibitem[{Sheng et~al.(2008)Sheng, Provost, and Ipeirotis}]{sheng2008get}
Victor~S Sheng, Foster Provost, and Panagiotis~G Ipeirotis. 2008.
\newblock Get another label? improving data quality and data mining using
  multiple, noisy labelers.
\newblock In \emph{Proceedings of the 14th ACM SIGKDD international conference
  on Knowledge discovery and data mining}, pages 614--622. ACM.

\bibitem[{Snow et~al.(2008)Snow, O'Connor, Jurafsky, and Ng}]{snow2008cheap}
Rion Snow, Brendan O'Connor, Daniel Jurafsky, and Andrew~Y Ng. 2008.
\newblock Cheap and fast---but is it good?: evaluating non-expert annotations
  for natural language tasks.
\newblock In \emph{Proceedings of the conference on empirical methods in
  natural language processing}, pages 254--263. Association for Computational
  Linguistics.

\bibitem[{Stenetorp et~al.(2012)Stenetorp, Pyysalo, Topi\'{c}, Ohta, Ananiadou,
  and Tsujii}]{brat}
Pontus Stenetorp, Sampo Pyysalo, Goran Topi\'{c}, Tomoko Ohta, Sophia
  Ananiadou, and Jun'ichi Tsujii. 2012.
\newblock \href {http://dl.acm.org/citation.cfm?id=2380921.2380942} {Brat: A
  web-based tool for nlp-assisted text annotation}.
\newblock In \emph{Proceedings of the Demonstrations at the 13th Conference of
  the European Chapter of the Association for Computational Linguistics}, EACL
  '12, pages 102--107, Stroudsburg, PA, USA. Association for Computational
  Linguistics.

\bibitem[{Vaswani et~al.(2017)Vaswani, Shazeer, Parmar, Uszkoreit, Jones,
  Gomez, Kaiser, and Polosukhin}]{transformer}
Ashish Vaswani, Noam Shazeer, Niki Parmar, Jakob Uszkoreit, Llion Jones,
  Aidan~N Gomez, \L~ukasz Kaiser, and Illia Polosukhin. 2017.
\newblock Attention is all you need.
\newblock In \emph{Proceedings of NIPS}. Curran Associates, Inc.

\bibitem[{Wallace et~al.(2011)Wallace, Small, Brodley, and
  Trikalinos}]{wallace2011should}
Byron~C Wallace, Kevin Small, Carla~E Brodley, and Thomas~A Trikalinos. 2011.
\newblock Who should label what? instance allocation in multiple expert active
  learning.
\newblock In \emph{Proceedings of the 2011 SIAM International Conference on
  Data Mining}, pages 176--187. SIAM.

\bibitem[{Yan et~al.(2011)Yan, Rosales, Fung, and Dy}]{yan2011active}
Yan Yan, Romer Rosales, Glenn Fung, and Jennifer~G Dy. 2011.
\newblock Active learning from crowds.
\newblock In \emph{ICML}, volume~11, pages 1161--1168.

\bibitem[{Zhai et~al.(2013)Zhai, Lingren, Deleger, Li, Kaiser, Stoutenborough,
  and Solti}]{zhai2013web}
Haijun Zhai, Todd Lingren, Louise Deleger, Qi~Li, Megan Kaiser, Laura
  Stoutenborough, and Imre Solti. 2013.
\newblock Web 2.0-based crowdsourcing for high-quality gold standard
  development in clinical natural language processing.
\newblock \emph{Journal of medical Internet research}, 15(4).

\end{thebibliography}
\bibliographystyle{acl_natbib}

\end{document}